\documentclass[10pt,twocolumn,letterpaper]{article}

\usepackage[pagenumbers]{cvpr} 

\usepackage{graphicx}
\usepackage{amsmath}
\usepackage{amssymb}
\usepackage{booktabs}

\usepackage[pagebackref,breaklinks,colorlinks]{hyperref}

\usepackage[capitalize]{cleveref}
\crefname{section}{Sec.}{Secs.}
\Crefname{section}{Section}{Sections}
\Crefname{table}{Table}{Tables}
\crefname{table}{Tab.}{Tabs.}

\def\cvprPaperID{7850} 
\def\confName{CVPR}
\def\confYear{2022}

\begin{document}

\title{Partially-Supervised Novel Object Captioning Leveraging Context from Paired Data}

\author{Shashank Bujimalla\\
Intel Corporation\\
{\tt\small shashankbvs@gmail.com}
\and
Mahesh Subedar\\
Intel Labs\\
{\tt\small mahesh.subedar@intel.com}
\and
Omesh Tickoo\\
Intel Labs\\
{\tt\small omesh.tickoo@intel.com}
}

\maketitle

\begin{abstract}
In this paper, we propose an approach to improve image captioning solution for images with novel objects that do not have caption labels in the training dataset. We refer to our approach as Partially-Supervised Novel Object Captioning (PS-NOC). PS-NOC is agnostic to model architecture, and primarily focuses on the training approach that uses existing fully paired image-caption data and the images with only the novel object detection labels (partially paired data). We create synthetic paired captioning data for novel objects by leveraging context from existing image-caption pairs. We then create pseudo-label captions for partially paired images with novel objects, and use this additional data to fine-tune the captioning model. We also propose a variant of SCST within PS-NOC, called SCST-F1, that directly optimizes the F1-score of novel objects. Using a popular captioning model (Up-Down) as baseline, PS-NOC sets new state-of-the-art results on \textit{held-out} MS COCO out-of-domain test split, i.e., 85.9 F1-score and 103.8 CIDEr. This is an improvement of 85.9 and 34.1 points respectively compared to baseline model that does not use partially paired data during training. We also perform detailed ablation studies to demonstrate the effectiveness of our approach.
\end{abstract}

\section{Introduction}
\label{sec:intro}

Image captioning solutions generate natural language description (i.e., caption) of the salient visual concepts in a given image. There has been a rapid progress in using deep neural network models to develop automatic image captioning solutions. The datasets that are used to train these models contain paired image-caption data, i.e., each image consists of a set of human annotated ground-truth captions; we also refer to such data as fully paired data from hereon. These datasets are often limited in the number and diversity of visual concepts; for e.g., MS COCO Captions~\cite{chen2015microsoft}, which is the most popular captioning dataset contains only 91 underlying object classes. This could severely hinder the practical application of such models. There has been an increased effort to improve image captioning solutions on novel object classes that do not have paired image-caption annotations in training dataset. These are referred to as Novel Object Captioning (NOC) solutions, and benchmark datasets such \textit{held-out} MS COCO~\cite{hendricks2016deep} and \textit{no-caps}~\cite{agrawal2019nocaps} have been created to evaluate them. 

\begin{figure}[!t]
\captionsetup[subfigure]{labelformat=empty}
		\centering     
		\small
        \begin{subfigure}[t]{0.9\columnwidth}
        \centering
		\includegraphics[width=0.85\textwidth, height=20mm]{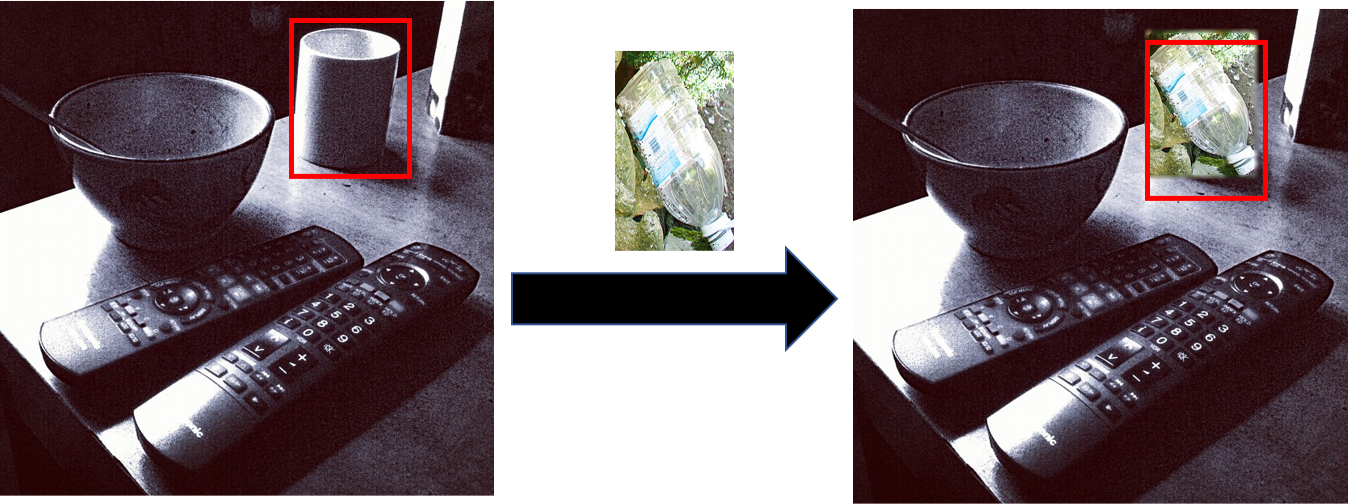}
		\end{subfigure}
		
	    \begin{subfigure}[t]{0.45\columnwidth}
	    \vspace{-2mm}
		\caption{A bowl a \textbf{\emph{cup}} and two controllers on a table.}
		\end{subfigure}
		\hspace{3mm}
		\begin{subfigure}[t]{0.45\columnwidth}
		\vspace{-2mm}
		\caption{A bowl a \textbf{\emph{bottle}} and two controllers on a table.}
		\end{subfigure}

		\begin{subfigure}[t]{0.25\columnwidth}
        \centering
		\includegraphics[width=1.00\columnwidth, height=20mm]{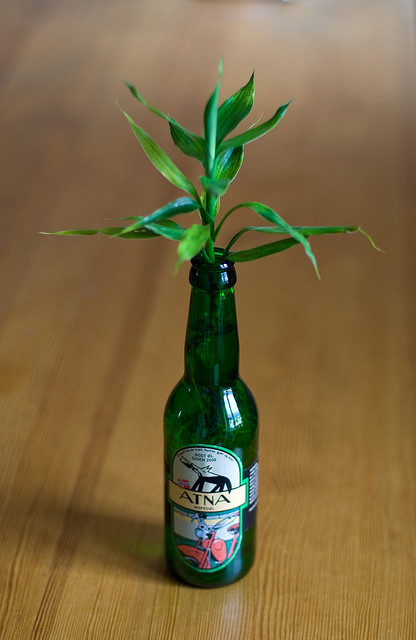}
		\end{subfigure}
		\hspace{2mm}
		\begin{subfigure}[t]{0.55\columnwidth}
        \centering
        \vspace{-20mm}
		\caption{\textbf{Baseline:} A green flower in a glass on a table.\\\\
		\textbf{Ours:} A \textbf{\underline{bottle}} with a flower in it on a table.}
		\end{subfigure}
	 
	\caption{\small In our PS-NOC approach, we generate synthetic paired data for novel objects, as shown in `bottle' example on the top. We propose a three-step training technique that uses this additional synthetic data and partially paired data to generate captions for novel objects. Our approach generates a caption that correctly includes `bottle' on an unseen image in this example. }
\label{fig:intro_aug_examples}	
\end{figure}

The current NOC solutions broadly belong to two categories. The first category of solutions use specialized model architectures to insert the objects detected in an image into the caption~\cite{hendricks2016deep,venugopalan2017captioning,chen2021leveraging,cagrandi2021learning,lu2018neural,yao2017incorporating,wang2021ecol}. The second category of solutions use partial supervision to train the models on images with novel objects~\cite{anderson2018partially, cao2020feature}.
The partially-supervised solutions use images from object detection datasets, which include a much wider variety of object classes than the captioning datasets, to train the captioning model. 
For e.g., Open Images~\cite{OpenImages2} contains 600 object classes that includes human annotations for object bounding boxes along with their class labels. These images are referred to as partially paired data from hereon, since their object class labels can be considered as partial ground-truth captions.
In this work, we focus on the second category of solutions and propose our Partially-Supervised Novel Object Captioning (PS-NOC) solution that is agnostic to model architecture.

We first generate synthetic paired image-caption data for novel objects. The current partially supervised solutions for NOC, i.e., PS3~\cite{anderson2018partially} and FDM-net~\cite{cao2020feature}, use the images and the class labels of partially paired data, but do not use their bounding box annotations. We use these bounding box annotations and generate synthetic images by replacing objects in fully paired images with novel objects. 
We also generate the corresponding synthetic captions for these synthetic images by direct word replacement of objects in the fully paired captions with novel objects. 
We refer to this new paired image-caption data as synthetic data, although they are extracted from natural image-caption pairs by leveraging the context of objects in fully-paired data.
Such synthetic images are often prone to noise~\cite{cao2020feature}; so, we use simple heuristics like restricting certain object replacements to reduce the noise. Similarly, we use simple heuristics to reduce the noise in synthetic captions.
An example of such synthetic data is shown in Figure~\ref{fig:intro_aug_examples}.

We then propose a novel three step training technique that uses this synthetic data and partially paired data to train the captioning models for novel objects.
In practical real world situations, there exist sophisticated captioning models that are pre-trained on fully paired data. Our first step mimics this by training a popular captioning model, Up-Down ~\cite{anderson2018bottom}, on fully paired data.
In the second step, we augment our fully paired data with synthetic data and warm up the captioning model to novel objects; we notice this considerably improves the captioning model performance on images with these novel objects.
Finally, in the third step, we use this model to iteratively generate pseudo-label captions for partially paired images and further fine-tune the model using this additional paired data.
Figure~\ref{fig:intro_aug_examples} shows an example novel object caption generated using our approach.

We use several novel techniques during our proposed PS-NOC training process. For each partially paired image, we generate two pseudo-label captions, one caption that prioritizes novel object inclusion and the other that prioritizes caption quality. We also use offline pseudo-labeling, i.e., we fine-tune the model for several thousands of mini-batch iterations and then generate pseudo-label captions. This makes our approach easy to integrate on existing implementations of captioning models. We also propose SCST-F1, a variant of Self-Critical Sequence Training (SCST)~\cite{rennie2017self}, for NOC that directly optimizes the F1-score of novel objects; the F1-score is a standard NOC evaluation metric~\cite{hendricks2016deep} along with CIDEr~\cite{vedantam2015cider}. This SCST formulation is possible with our training approach since we have access to training data containing novel objects, i.e. synthetic paired data and pseudo-labelled partially paired data.

In summary, our main contributions are as follows:

\textbf{1.}  We generate synthetic paired image-caption data for novel objects by using object bounding box annotations and leveraging the context from fully paired data. To our knowledge, our work is the first to study the usage of such synthetic paired data with images for NOC.

\textbf{2.} We propose a model-agnostic three step training approach that uses the additional synthetic data and partially paired data to train the captioning model. This includes two novel training techniques: i) SCST-F1, which directly optimizes F1-score of novel objects, ii) Pseudo-labeling for NOC, which makes use of partially paired images, considers novel object inclusion along with caption quality, and is easy to integrate into existing implementations. 
Using Up-Down captioning model as baseline, our PS-NOC approach sets new state-of-the-art results on \textit{held-out} MS COCO out-of-domain (i.e., novel object images) test split with 85.9 F1-score and 103.8 CIDEr.

\section{Related Work}

Many of the current solutions for NOC use specialized model architectures to generate captions that contain novel objects. For e.g., they use two parallel network pipelines ~\cite{hendricks2016deep,venugopalan2017captioning,chen2021leveraging,cagrandi2021learning}, neural slot-filling ~\cite{lu2018neural}, or copying mechanism  ~\cite{yao2017incorporating,wang2021ecol}.
There also exist a few partially-supervised solutions, PS3~\cite{anderson2018partially} and FDM-net~\cite{cao2020feature}, that use images from object detection datasets, i.e., partially paired images, to train the captioning model.

PS3~\cite{anderson2018partially} uses constrained beam search (CBS)~\cite{anderson2016guided} to generate pseudo-label captions for the partially paired images. CBS forces the inclusion of specified tag words, called constraints, into the model generated captions. To generate these tag words for partially paired images, PS3 uses an additionally trained lexical classifier from~\cite{hendricks2016deep}. PS3 generates these pseudo-label captions at each mini-batch of training and refer to it as online version. To perform this training in a compute-efficient manner, they maintain two copies of the model with tied weights. 
In our PS-NOC approach, we perform offline version of pseudo-labeling by generating pseudo labels after training the model for several thousands of mini-batch iterations. Since we use synthetic data to warm-up the model to novel objects before performing pseudo-labeling, we notice that our model fine-tuning converges within a few rounds (for e.g., 1 to 4) of offline pseudo-labeling. This makes our approach easier to integrate on any existing captioning model implementations, as we do not need two copies of our model during training. However, our approach is also applicable to online version, if required. Unlike PS3, we do not use the additional lexical classifier and instead use the object detection labels directly as CBS constraints. In PS-NOC, we also generate an additional pseudo-label caption that prioritizes caption quality over novel object inclusion. 

FDM-net~\cite{cao2020feature} uses partially paired images and generates synthetic data containing novel objects. However, they perform object replacement in image feature space and generate synthetic region features by using object detection model. This restricts FDM-net's application to region based captioning models. Our PS-NOC approach to use synthetic images decouples this dependency, making it more generic and applicable to even the advancing captioning models that use the entire image~\cite{wang2021simvlm, liu2021cptr}. Our approach also allows us to include humans in the loop to sign-off or discard the synthetic image-caption data generated using such automation. 
FDM-net also generates synthetic captions, but trains an additional scene-graph based network to generate them. We instead generate synthetic captions by direct word replacement, and then use simple heuristics to remove noise in these captions. 

\begin{figure}[!t]
\captionsetup[subfigure]{labelformat=empty}
		\centering     
		\small
        \begin{subfigure}[t]{0.9\columnwidth}
        \centering
		\includegraphics[width=1.0\columnwidth]{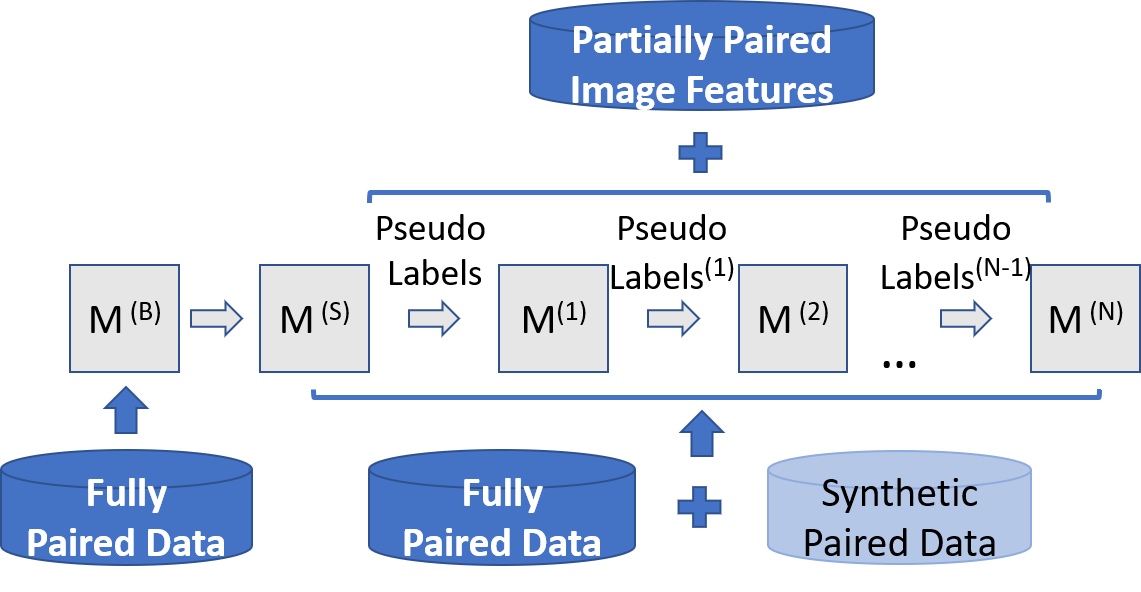}
		\end{subfigure}
	\caption{Overview of our training technique.}
\label{fig:training_ovw}	
\end{figure}

\begin{figure*}[!t]
\captionsetup[subfigure]{labelformat=empty}
		\centering     
	    \begin{subfigure}[t]{0.47\textwidth}
		\includegraphics[width=0.95\columnwidth, height=22mm]{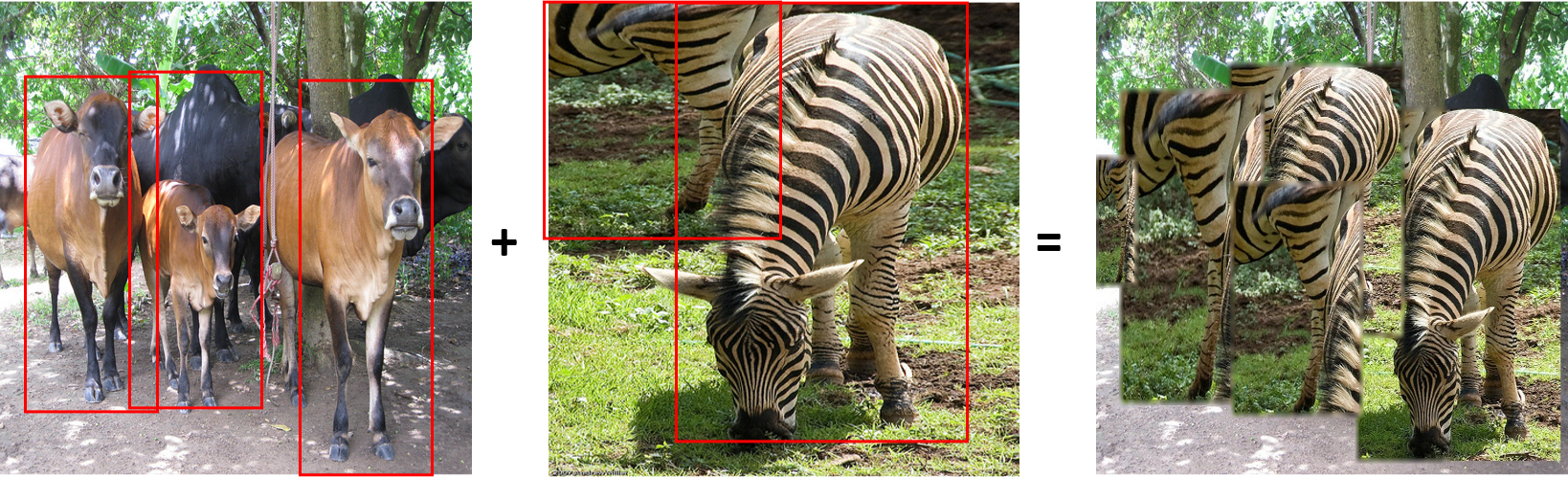}
		\end{subfigure}
		\hspace{2mm}
			\begin{subfigure}[t]{0.47\textwidth}
    	\vspace{-18mm}
		\caption{A group of cows on dirt area with trees in background.\\+ Novel Object: zebra\\= A group of zebras on dirt area with trees in background.
		}
		\end{subfigure}
		\vspace{2mm}
		
		 \begin{subfigure}[t]{0.47\textwidth}
		\includegraphics[width=0.95\columnwidth, height=22mm]{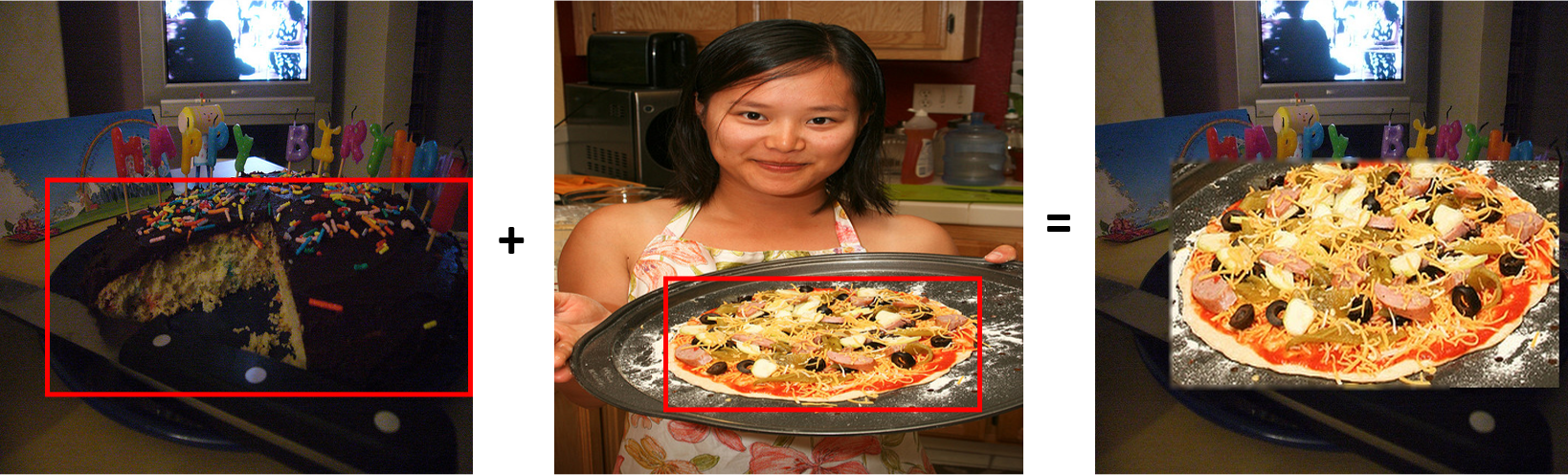}
		\end{subfigure}
			\hspace{2mm}
			\begin{subfigure}[t]{0.47\textwidth}
    	\vspace{-23mm}
		\caption{A blue plate holding a frosted cake and knife.\\+ Novel Object: pizza\\= A plate holding a pizza and knife. \\\\
		A birthday cake has a fraction of itself cut and eaten.\\+ Novel Object: pizza\\= A pizza has a fraction of itself cut and eaten.
		}
		\end{subfigure}
		

	\caption{Examples of synthetic data generated using our approach. The first two captions are accurate while the third has incorrect context (`pizza' is not cut or eaten).}
\label{fig:aug_examples}	
\end{figure*}

\section{Our Approach: PS-NOC}


\subsection{Synthetic Data Generation}

We first generate paired synthetic data, i.e., image-caption pairs, for novel objects. To generate synthetic images, we use the bounding box annotations of novel objects in partially paired images and replace them into bounding box annotations of candidate objects in fully paired images. We also generate the corresponding synthetic captions for these synthetic images by directly replacing the objects in fully paired captions with novel objects. Examples of synthetic data generated using our approach are shown in Figure~\ref{fig:aug_examples}. More details are as follows. 

\textbf{Candidate object selection}: We follow the approach similar to FDM-net~\cite{cao2020feature} to identify the candidate objects in fully paired data for replacement with a novel object. In this approach, validation split images containing novel objects are input to a baseline captioning model that is trained on fully paired data. For each novel object, the number of occurrences of each fully paired object in the generated captions is used to identify the $m$ most likely candidate objects that the model finds similar to the novel object. 
We generate up to $K$ synthetic images for each novel object, with $K/m$ images for each novel-candidate object pair. We also heuristically try to minimize the number of times each novel object image is used, so that we generate synthetic images with diverse novel object features and background contexts. 

\textbf{Image selection}: The sizes of novel objects and candidate objects could vary widely across images. In order to avoid a big change in resolution and aspect ratio when performing object replacement, we only consider images where novel objects and candidate objects have bounding box annotations whose areas do not differ by more than $\Delta A_{max}\%$; we also set a lower limit, $A_{min}$, on their areas. We set similar constraints ${AR}_{min}$ and $\Delta{AR}_{max}\%$ for their aspect ratios, along with an upper limit on aspect ratio ${AR}_{max}$. 
If multiple occurrences of an object are found in each image, we consider all the pairs for replacement and exclude the pairs that do not meet the constraints. For e.g., multiple occurrences of `cow' are replaced with multiple occurrences of `zebra' in Figure~\ref{fig:aug_examples}. We observe that including such images with multiple object occurrences into synthetic data is critical for captioning performance. 
We also exclude any objects whose bounding box annotations completely contain other annotations; this is to avoid cases where additional objects accidentally sneak into the synthetic image along with novel objects, or get removed along with candidate object, thus resulting in a noisy caption. 

\textbf{Synthetic caption processing}: When candidate objects get replaced with novel objects, their descriptions in the caption could be incorrect. 
We use simple heuristics to process these captions. Before the candidate object is replaced with novel object, we remove any occurence of color words (e.g., `blue') in the caption; we also remove any adjectives or nouns that occur within certain word radius, $\Delta R_{adj}$ and $\Delta R_{noun}$ respectively, of the candidate object word. For e.g., `a frosted cake' is converted to `a pizza' in Figure~\ref{fig:aug_examples}. 

We notice that synthetic images and captions generated using our simple heuristics are effective enough to improve the overall NOC results, as shown in Section~\ref{sec:experiments}.


\subsection{Training Technique}
\label{sec:training_tech}

Figure~\ref{fig:training_ovw} shows the overview of our training technique that consists of three steps, and leverages synthetic and partially paired data. 

\textbf{Step I. Train using fully paired data}: We first train our baseline captioning model $\text{M}^{\text{(B)}}$ with fully paired data. 

\textbf{Step II. Fine-tune using synthetic data}: We warm-up our captioning model $\text{M}^{\text{(S)}}$ to novel objects using our synthetic data in addition to the fully paired data. 

\textbf{Step III. Generate pseudo-labels on partially paired data and fine-tune}: In this step, we use our model from Step 2 that considerably improved its generalization to novel objects, and generate pseudo-label captions for partially paired images that contain novel objects. This additional partially paired data, on top of fully paired and synthetic data, is used to further fine-tune our model. We repeat Step III a fixed number of times, i.e., $N$ rounds, to generate the final model $\text{M}^{\text{(N)}}$. 

\textbf{Pseudo-labeling approach}: 
Our pseudo-labeling approach in Step III consists of two key components. 1. We perform offline pseudo-labeling, i.e., we fine-tune the model for several thousands of mini-batch iterations and generate pseudo-labels. This makes it easy to integrate our approach on any existing implementations of captioning models.
2. For each partially paired image, we generate two pseudo-label captions using beam search: one caption using CBS~\cite{anderson2016guided}, and the other caption without using CBS. We observe that CBS could sometimes negatively impact caption quality (i.e., CIDEr~\cite{vedantam2015cider} score) while including the novel objects in the caption. On the other hand, we notice that disabling CBS improves the caption quality while missing some novel objects due to the limitations in synthetic data. Using both these captions as pseudo-labels on top of our approach outperforms using CBS caption alone. We provide this ablation study results in Supplemental material.

\subsection{SCST-F1}

Image captioning models are initially trained using cross-entropy loss, and then using Self-Critical Sequence Training (SCST)~\cite{rennie2017self,bujimalla2020b,Cornia_2020_CVPR}. SCST uses policy gradient based RL to directly optimize CIDEr~\cite{vedantam2015cider}, which is a widely accepted captioning quality metric.
In general, a NOC dataset is divided into in-domain and out-of-domain splits, wherein the out-of-domain designation is given to an image if there is a mention of any of the novel objects in at least one of its ground-truth captions. The F1-score of novel objects in out-of-domain split~\cite{hendricks2016deep} captions is a measure of model's ability to accurately include novel objects in predicted captions. It is a widely accepted NOC metric along with out-of-domain CIDEr.

We propose a variant of SCST that directly optimizes F1-score, and refer to it as SCST-F1. Note that our approach facilitates this formulation since we have access to out-of-domain training data, i.e., either synthetic images with their synthetic captions, partially paired images with their pseudo-label captions, or both. 
We split the SCST reward into three components: in-domain CIDEr, out-of-domain CIDEr and out-of-domain F1-score; these are denoted using $r_1(.)$, $r_2(.)$ and $r_3(.)$ respectively.
We use beam search with beam size $k$ during SCST and track top-$k$ most probable captions for each sample image. For each reward component $r_j(.)$, the mean score of these top-$k$ captions~\cite{Cornia_2020_CVPR} is used as its RL baseline $\tilde{r_j}$ during SCST loss calculation. The gradient of SCST-F1 loss for each sample is approximated as:
\[
\begin{aligned}
\nabla_{\theta}L(\theta) {} \approx -\frac{1}{k}\sum_{i=1}^k({} & \alpha_1({r_1}({y^i}) - \tilde{r_1}) + \alpha_2({r_2}({y^i}) - \tilde{r_2}) + \\
& \alpha_3({r_3}({y^i}) - \tilde{r_3}))\nabla_{\theta}\log p_{\theta}({y^i}),  \\
\end{aligned}
\]
where $\theta$ denotes the model parameters, $y^i$ is the caption in $i$-th beam, and $\tilde{r_j} = \frac{1}{k}\sum_{i=1}^k{r_j}({y^i})$ for $j = 1, 2, 3$. We set $r_1(y^i)=0$ for an out-of-domain sample, and $r_2(y^i)=0$, $r_3(y^i)=0$ for an in-domain sample. The hyper-parameters $\alpha_1$, $\alpha_2$, $\alpha_3$ are the weights given to each of the three reward components. 

\subsection{Training Loss}
We perform training using only the cross-entropy loss in Step I.
In Step II, we first perform training using cross-entropy loss and then perform SCST-F1. In Step III, we directly perform SCST-F1.  

\section{Experiments}
\label{sec:experiments}


%
%

\subsection{Dataset} 
We evaluate our approach on \textit{held-out} MS COCO dataset, using the splits provided by DCC~\cite{hendricks2016deep} for NOC. This dataset is a subset of 2014 MS COCO~\cite{chen2015microsoft} that originally contains training split with 82,783 images and validation split with 40,504 images. The \textit{held-out} dataset excludes all image-caption pairs that describe any of the eight selected objects (`bottle', `bus', `couch', `microwave', `pizza', `racket', `suitcase', , `zebra') from training split, which results in a training split that contains 70,194 images with fully paired data. These eight object classes are referred to as novel object classes from hereon. In our trials, we consider the remaining 12,589 images excluded from the training split as partially paired data.
DCC also provides \textit{held-out} validation and test splits that each contain 20,252 images; they further divide each of these splits into in-domain and out-of-domain splits for NOC evaluation. 

\begin{table*}[!t]
\caption{\small Results of our PS-NOC approach on test-split and comparison against state-of-the-art results. PS-NOC provides the highest scores for both out-of-domain CIDEr and F1-scores, which is also seen in their COF1 and COF1.5 scores. Soln-2 includes additional post-processing during SCST-F1 training and inference to improve the caption quality. 
} 
\vspace{-4mm}
\label{tab:results}
\begin{center}
\begin{small}
\begin{tabular}{ccccccccccccc}
\toprule
& & & \multicolumn{6}{c}{Out-of-domain} && \multicolumn{3}{c}{In-domain} \\\cmidrule{4-9}\cmidrule{11-13} 
ID & Approach & C-RL &  S &  M &  C &  F1 &   COF1 &   COF1.5 &&  S &  M &  C \\\midrule
1a & PS-NOC (Soln-1) & Yes & \textbf{19.7} & \textbf{27.2} & \textbf{101.5} & \textbf{86.1} & \textbf{93.2} & \textbf{96.2} && 19.2 & 26.9 & 110.1 \\
1b & PS-NOC (Soln-2) & Yes & \textbf{20.8} & \textbf{28.0} & \textbf{103.8} & \textbf{85.9} & \textbf{94.0} & \textbf{97.6} && 20.5 & 27.7 & 110.9 \\
2 & PS3 & No & 17.9 & 25.4 & 94.5 & 63 & 75.6 & 81.9 && 19.0 & 25.9 & 101.1 \\
3a & FDM (no CBS) & No & 19.4 & 25.9 & 84.8 & 64.7 & 73.4 & 77.4 && 20.2 & 27.2 & 109.7 \\
3b & FDM (CBS) & No & 19.6 & 25.6 & 85.3 & 85.7 & 85.5 & 85.4 && 19.7 & 26.2 & 105.5 \\
4 & NBT (CBS) & No & 17.4 & 24.1 & 86.0 & 70.3 & 77.4 & 80.5&& 18.0 & 25.0 & 92.1 \\
5a & Region Selector & No & 18.3 & 24.9 & 78.2 & 75.0 & 76.6 & 77.2 && 19.2 & 26.2 & 97.0 \\
5b & Region Selector (DGBS) & Yes & 19.4 & 26.3 & 88.5 & 75.1 & 81.3 & 83.9&& 21.0 & 27.9 & 115.3 \\
6 & ANOC & Yes & 18.2 & 25.2 & 94.7 & 64.3 & 76.6 & 82.7 && - & - & - \\
7 & ECOL-R (CBS) & Yes & 19.1 & 25.7 & 99.1 & 71.8 & 83.3 & 88.7 && 20.8 & 26.8 & 112.6 \\
 \bottomrule
\end{tabular}
\end{small}
\end{center}
\end{table*}

\begin{table*}[!t]
\caption{\small Results of our PS-NOC approach on test split by setting different values of $\alpha_3$ during our SCST-F1 in Step II. We train our model using only Steps I and II for these trials. The F1-score increases as $\alpha_3$ increases demonstrating that our SCST-F1 technique is effective.} 
\vspace{-4mm}
\label{tab:scstf1}
\begin{center}
\begin{small}
\begin{tabular}{ccccccccc}
\toprule
& \multicolumn{3}{c}{Weight factors} && In-domain  && \multicolumn{2}{c} {Out-of-domain}\\ \cmidrule{2-4} \cmidrule{6-6} \cmidrule{8-9} 
Row & $\alpha_1$ &  $\alpha_2$ & $\alpha_3$ &&  C &&  C &  F1  \\\midrule
1 & 1 &  1 &  0 &&  113.4 &&	98.6 &	70.8 \\
2 & 1 & 1 & 1 && 111.3 && 99.2 & 76.4 \\
3 & 1 & 1 & 5 && 108.9 && 99.5 & 81.7 \\
4 & 1 & 1 & 10 && 107.4 && 98.7 & 82.4 \\
5 & 1 & 1 & 20 && 104.4 && 97.3 & 84.8 \\
\bottomrule
\end{tabular}
\end{small}
\end{center}
\end{table*}

\subsection{Implementation Details} 
\label{sec:imp_details}

\hspace{\parindent}\textbf{Synthetic data generation}: To generate synthetic data for each novel object, we use the top similar candidate objects in the fully paired data that were identified by FDM-net~\cite{cao2020feature}. For e.g., fully paired images with `cup', `glass', or `vase' are considered as candidates to generate synthetic data for novel object `bottle'. We exclude four of these candidate object classes (`stove', `bread', `box', and `racquet') as they don't have bounding box annotations in MS COCO dataset. We show our final list in Supplemental material. We generate at most $K=2400$ synthetic images for each novel object; if there are $m=3$ candidate objects for a novel object (for e.g., `bottle' has `cup', `glass', `vase' candidates), we generate at most $K/m = 800$ images for each novel-candidate object pair (for e.g., `bottle',`cup').
We set $A_{min}=1000$ pixels, $\Delta A_{max}=200\%$, ${AR}_{min}=0.05$, ${AR}_{max}=5.0$, and $\Delta {AR}_{max}=30\%$.  
To generate synthetic captions, we use NLTK~\cite{10.3115/1118108.1118117} part-of-speech tagger to identify adjectives and nouns; we set $\Delta {R}_{adj} = 2$, and $\Delta {R}_{noun} = 1$. 
These hyper-parameters were empirically selected by manually inspecting the quality of few synthetic images and their captions, and are by no means the most optimal values.

\textbf{Model Architecture}: We use the popular Bottom-Up Top-Down (Up-Down) captioning model~\cite{anderson2018bottom} that is based on encoder-decoder neural architecture. The encoder consists of a Faster R-CNN~\cite{ren2015faster} object detector that is trained on Visual Genome dataset~\cite{krishna2017visual}. For each image $I$, a set of feature vectors $V$ associated with salient bounding boxes in the image are extracted using Faster R-CNN. At each timestep~$t$ (i.e., word in the caption), the decoder, which consists of a two-layer Long Short-Term Memory (LSTM) network, uses~$V$ and previous word~$y_{t-1}$ in the caption to compute the conditional probability of the next word $p_{\theta}(y_t\,|\,y_{1:t-1}, I)$ for all the words in the word vocabulary, where~$\theta$ are the model parameters. 
We use Faster R-CNN with ResNext-152 as backbone. In the Supplemental material, we give details on the reason for choosing this backbone and why it does not provide us an unfair advantage during comparison with previous works.
For language embeddings, we follow the same approach as PS3~\cite{anderson2018partially}, and add pre-trained word embeddings to the input and output layers of the decoder. 
However, we simply use GloVe embeddings~\cite{pennington2014glove} and freeze them throughout our model training. 

\textbf{Object Detection Labels and CBS}: We use the Faster R-CNN model trained on Open Images from \textit{no-caps} baseline~\cite{agrawal2019nocaps} to generate object detection labels for the partially paired images. CBS in Step III of our model training uses these labels to generate one pseudo-label caption for each of these images.
We use the CBS implementation from \textit{no-caps} baseline~\cite{agrawal2019nocaps}. 

\textbf{Evaluation Metrics}: We use SPICE~\cite{anderson2016spice}, METEOR~\cite{denkowski2014meteor} and CIDEr~\cite{vedantam2015cider}  metrics to evaluate the caption quality; these are denoted as `S', `M' and `C' in our Results Tables. 
On out-of-domain test set, we also evaluate the F1-score metric of novel objects, denoted as `F1' in our Results tables. All the metric scores on out-of-domain test set are macro-averaged across the eight novel object classes for consistency with results reported in previous works~\cite{anderson2018partially}. 
We primarily focus on out-of-domain CIDEr and F1-score for comparison with state-of-the-art results as they are widely accepted as the caption fluency and NOC metrics respectively.
\textbf{F\textsubscript{\boldmath{$\beta$}}}\textbf{-score of CIDEr and F1-score}: 
We often notice a trade-off between out-of-domain CIDEr and F1-score in state-of-the-art results (see Table~\ref{tab:results}), which makes the comparison of results non-trivial. In order to represent them into a single metric for comparison, we also measure the  F\textsubscript{$\beta$}-score~\cite{powers2011evaluation} of out-of-domain CIDEr and F1-score with $\beta$ deciding the relative importance of CIDEr. $\beta=1$ gives equal importance to both metrics, while $\beta>1$ gives more importance to caption fluency (i.e, CIDEr) than including the novel objects in caption and degrading its fluency. We evaluate the models with $\beta$ set to 1 and 1.5, and denote them as `COF1' and `COF1.5' in our Results Tables. 

\textbf{Training Details}:
For all the steps in our model training, we use a learning rate (LR) schedule that linearly decreases from initial rate to 0. We use a batch size of 100 for training with cross-entropy loss. During SCST-F1 training, we use a batch size of 50, beam search with beam size $k=5$ and set $\alpha_1=1$, $\alpha_2=1$, $\alpha_3=1$. 
In \textbf{Step I}, we train our baseline model for 40,000 iterations with an initial LR of 0.01.
In \textbf{Step II}, we first fine-tune our model with cross-entropy loss for 15,000 iterations with initial LR of 0.005. We then fine-tune it using SCST-F1 for 10,000 iterations with initial LR of 0.0125.
In \textbf{Step III}, we further fine-tune our model using SCST-F1 for $N=4$ rounds with 6,000 iterations in each round. We use an initial LR of 0.002 for first round, and scale it by 0.8 across rounds.
We performed manual hyper-parameter tuning using CIDEr and F1-scores of validation split to set these hyper-parameter values; so, there may be scope to further improve these settings.

\textbf{Model Checkpoints}:
During Steps II and III of our training, we use the validation set scores to save the model checkpoint with the highest out-of-domain CIDEr, while also considering F1-score. Specifically, we allow a slight drop (1 point) in out-of-domain CIDEr if the corresponding F1-score is higher. 

\textbf{Inference Details}: We use beam search with beam size $k=5$. We do not use CBS during inference to generate the final captions from our PS-NOC.

\textbf{Caption Post-processing}: We notice that using SCST during training sometimes results in captions that end with the words `in', `a', `with', etc. To improve caption quality, we perform post-processing during SCST-F1 training and inference, and remove such words at the end of the caption. We refer to our approach without such post-processing as Soln-1, and the one with the post-processing as Soln-2.
We used slightly different training schedule for PS-NOC Step III in our Soln-2, compared to Soln-1; these details are provided in Supplemental material.  


\begin{table*}[!t]
\caption{\small Test split results from the ablation studies on different components of our PS-NOC approach. The results demonstrate the benefits of using i. Synthetic data, ii. Pseudo-labeling, iii. SCST, iv. SCST-F1, and v. Our overall approach.} 
\vspace{-4mm}
\label{tab:ablations_diff_steps}
\begin{center}
\begin{small}
\begin{tabular}{cccccccccccccc}
\toprule
& & &\multicolumn{4}{c}{Out-of-domain} & & \multicolumn{3}{c}{In-domain}  \\ \cmidrule{4-7}\cmidrule{9-11} 
Row & C-RL & Training technique &  S &  M &  C &  F1 &&  S &  M &  C  \\\midrule
1 & No & I & 19.6 & 28.0 & 69.7 & 0.0 && 19.4 & 27.9 & 108.0 \\
2 & No & I + CBSInf & 18.1 & 26.1 & 76.6 & 56.2 && 17.6 & 25.8 & 88.8 \\
3 & No & I + II & 19.8 & 28.2 & 89.0 & 62.1 && 19.3 & 27.6 & 103.4 \\
4 & No & I + II + III & 19.9 & 28.4 & 96.3 & 75.8 && 19.5 & 27.8 & 105.9 \\
5 & Yes & I + II(SCST) & 20.5 & 27.8 & 98.6 & 70.8 && 20.2 & 27.6 & 113.4 \\
6 & Yes & I + II(SCST) + III & 20.2 & 28.3 & 99.8 & 72.4 && 19.9 & 27.9 & 108.2 \\
7 & Yes & I + II(SCST) + III(SCST) & 20.1 & 27.8 & 101.0 & 78.8 && 19.6 & 27.1 & 111.0 \\
8 & Yes & I + II(SCST-F1) & 20.6 & 28.1 & 99.2 & 76.4 && 19.9 & 27.6 & 111.3 \\
9 & Yes & I + II(SCST-F1) + III & 20.2 & 28.5 & 102.2 & 75.7 && 20.0 & 28.2 & 108.2 \\
10 & Yes & I + II(SCST-F1) + III(SCST-F1) Soln-1 & 19.7 & 27.2 & 101.5 & 86.1 && 19.2 & 26.9 & 110.1 \\
11 & Yes & I + II(SCST-F1) + III(SCST-F1) Soln-2 & 20.8 & 28.0 & 103.8 & 85.9 && 20.5 & 27.7 & 110.9 \\

 \bottomrule
\end{tabular}
\end{small}
\end{center}
\end{table*}

\subsection{Results}
\label{sec:results}

\textbf{Comparison with previous works}: In Table~\ref{tab:results}, we compare our PS-NOC results on the test split with previous works and state-of-the-art results.
The column C-RL indicates if the approach uses any form of CIDEr based RL. 
For PS-NOC, we provide the results for both Soln-1, Soln-2 (discussed in Section~\ref{sec:imp_details}). Soln-1 does not include any caption post-processing, while Soln-2 includes it.

PS-NOC results (Rows 1a, 1b) consistently set new state-of-the-art results for both out-of-domain CIDEr and F1-score. They are also the most balanced in terms of CIDEr and F1, which is reflected in their highest F\textsubscript{$\beta$}-scores in columns COF1, COF1.5. 
Our results outperform the FDM-net and PS3 results, which are the existing partially supervised solutions for this dataset. Also, PS-NOC does not use the lexical constraints (471 most common adjectives, verbs, noun forms generated by additional training of MS COCO) that PS3 uses during CBS. 


\textbf{Effectiveness of SCST-F1}: In Table~\ref{tab:scstf1}, we demonstrate that our proposed SCST-F1 is effective by training Step-II using different values of $\alpha_3$. We train each of these models using our PS-NOC Steps I and II, and evaluate the results on test split. Row 1 with $\alpha_3=0$ corresponds to regular CIDEr based SCST with no F1-score reward. As we increase $\alpha_3$ from 0 to 20, we see the corresponding increase in out-of-domain F1-score of test split. Note that for a given captioning model, there could be trade-off between in-domain CIDEr and out-of-domain CIDEr or F1-score. So, the values of $\alpha_1$, $\alpha_2$, $\alpha_3$ should be chosen accordingly. We give equal importance to all the three scores in this work and therefore set all these three values to 1. 

\begin{figure*}[!ht]
\captionsetup[subfigure]{labelformat=empty}
		\centering     
		
		\begin{subfigure}[t]{0.23\textwidth}
		\caption{zebra}
		\end{subfigure}
		\hspace{1mm}
		\begin{subfigure}[t]{0.23\textwidth}
		\caption{bus}
		\end{subfigure}
		\hspace{1mm}
		\begin{subfigure}[t]{0.23\textwidth}
		\caption{couch}
		\end{subfigure}
		\hspace{1mm}
		\begin{subfigure}[t]{0.23\textwidth}
		\caption{microwave}
		\end{subfigure}		
		
		\begin{subfigure}[t]{0.23\textwidth}
		
		\includegraphics[width=1.0\textwidth, height=28mm]{figs/cap_in_images_new/zebra_COCO_val2014_000000235642.jpg}
		\caption{\textbf{Baseline}: A group of animals that are standing in the grass. }		
	
		\end{subfigure}
		\hspace{1mm}
		\begin{subfigure}[t]{0.23\textwidth}
		\includegraphics[width=1.0\textwidth, height=28mm]{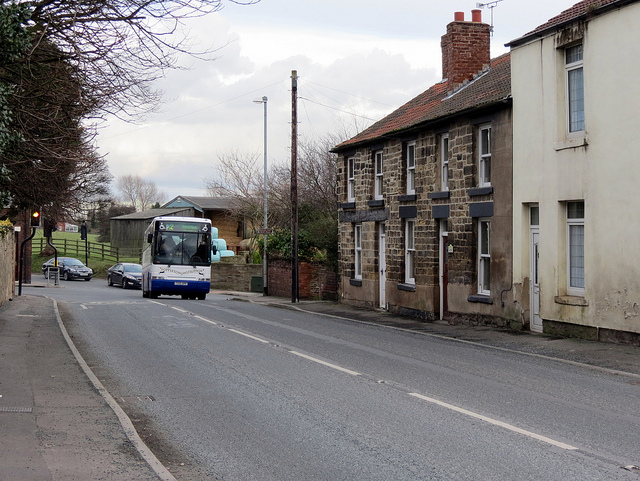}
		\caption{\textbf{Baseline}: A blue and white car on a city street.  
		}
		\end{subfigure}
		\hspace{1mm}
		\begin{subfigure}[t]{0.23\textwidth}
		\includegraphics[width=1.0\textwidth, height=28mm]{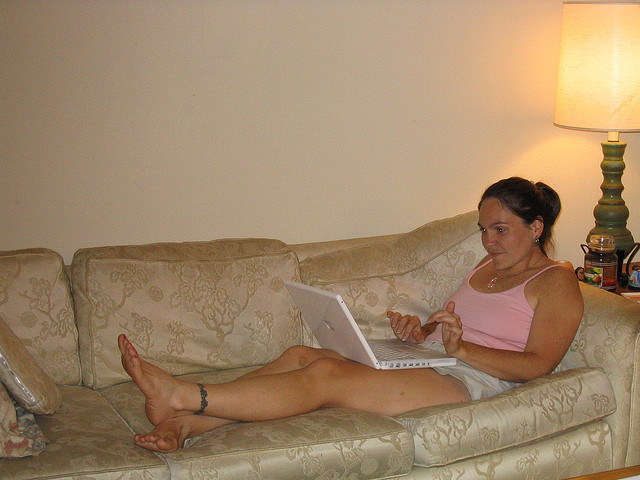}
		\caption{\textbf{Baseline}: A woman laying on a bed with a laptop.	}
		
		\end{subfigure}
		\hspace{1mm}
		\begin{subfigure}[t]{0.23\textwidth}
		\includegraphics[width=1.0\textwidth, height=28mm]{figs/cap_in_images_new/microwave_COCO_val2014_000000444275.jpg}
				\caption{
		\textbf{Baseline}: A tv sitting on the side of a road.
		}
		\end{subfigure}

		\begin{subfigure}[t]{0.23\textwidth}
		\caption{
		\textbf{Ours}: Group of \textbf{\underline{zebras}} are standing in a field.}
		\end{subfigure}
		\hspace{1mm}
		\begin{subfigure}[t]{0.23\textwidth}
		\caption{
		\textbf{Ours}: A \textbf{\underline{bus}} driving down a city street with cars.}
		\end{subfigure}
		\hspace{1mm}
		\begin{subfigure}[t]{0.23\textwidth}
		\caption{\textbf{Ours}: A woman sitting on a \textbf{\underline{couch}} using a laptop computer.}
		\end{subfigure}
		\hspace{1mm}
		\begin{subfigure}[t]{0.23\textwidth}
		\caption{
		\textbf{Ours}: A \textbf{\underline{microwave}} on the side of a road.}
		\end{subfigure}

		\vspace{2mm}
		
		\begin{subfigure}[t]{0.23\textwidth}
		\caption{pizza}
		\end{subfigure}
		\hspace{1mm}
		\begin{subfigure}[t]{0.23\textwidth}
		\caption{racket}
		\end{subfigure}
		\hspace{1mm}
		\begin{subfigure}[t]{0.23\textwidth}
		\caption{suitcase}
		\end{subfigure}
		\hspace{1mm}
		\begin{subfigure}[t]{0.23\textwidth}
		\caption{bottle}
		\end{subfigure}
		
		\begin{subfigure}[t]{0.23\textwidth}
		\includegraphics[width=1.0\textwidth, height=28mm]{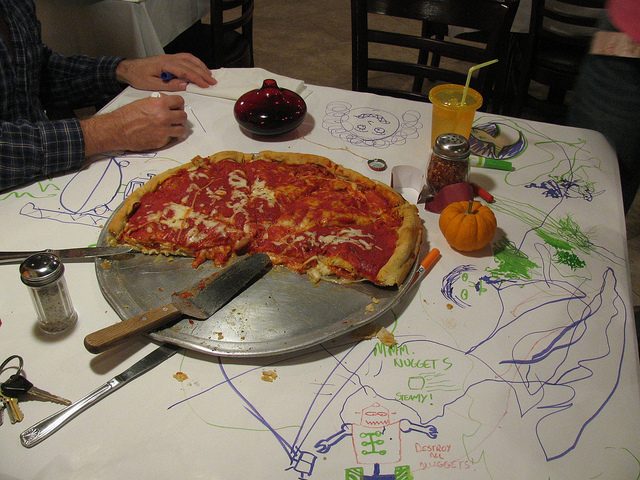}
		\caption{\textbf{Baseline}:  A man is cutting a large pie on a plate. }		
	
		\end{subfigure}
		\hspace{1mm}
		\begin{subfigure}[t]{0.23\textwidth}
		\includegraphics[width=1.0\textwidth, height=28mm]{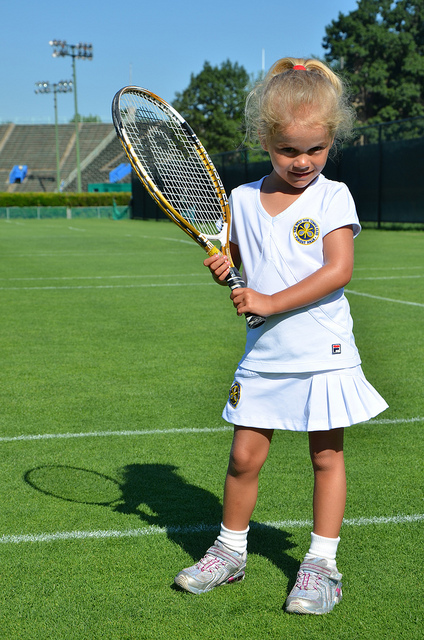}
		\caption{\textbf{Baseline}: A little girl standing on top of a lush green field. 
		}
		\end{subfigure}
		\hspace{1mm}
		\begin{subfigure}[t]{0.23\textwidth}
		\includegraphics[width=1.0\textwidth, height=28mm]{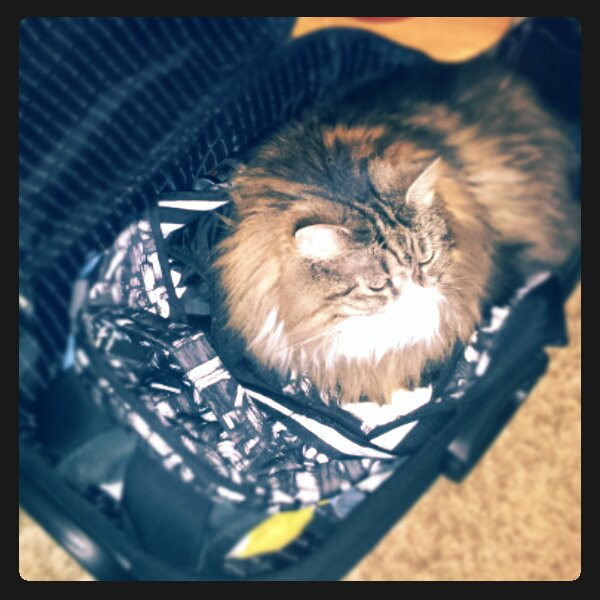}
		\caption{\textbf{Baseline}: A cat is sitting in a blue bowl.}
		
		\end{subfigure}
		\hspace{1mm}
		\begin{subfigure}[t]{0.23\textwidth}
		\includegraphics[width=1.0\textwidth, height=28mm]{figs/cap_in_images_new/bottle_COCO_val2014_000000580608.jpg}
				\caption{
		\textbf{Baseline}: A green flower in a glass on a table.
		}
		\end{subfigure}
		
		\begin{subfigure}[t]{0.23\textwidth}
		\caption{
		\textbf{Ours}: A \textbf{\underline{pizza}} on a plate with a knife.}
		\end{subfigure}
		\hspace{1mm}
		\begin{subfigure}[t]{0.23\textwidth}
		\caption{
		\textbf{Ours}: A little girl holding a tennis \textbf{\underline{racket}}  on a field.}
		\end{subfigure}
		\hspace{1mm}
		\begin{subfigure}[t]{0.23\textwidth}
		\caption{\textbf{Ours}: A cat sitting in a \textbf{\underline{suitcase}} on a floor. }
		\end{subfigure}
		\hspace{1mm}
		\begin{subfigure}[t]{0.23\textwidth}
		\caption{
		\textbf{Ours}: A \textbf{\underline{bottle}} with a flower in it on a table.}
		\end{subfigure}
		\caption{Examples of captions generated using our approach compared to the baseline model for the eight novel object classes.}
\label{fig:cap_examples}	
\end{figure*}

\textbf{Qualitative examples}:  We show qualitative examples of the captions generated using our PS-NOC Soln-1 in Figure~\ref{fig:cap_examples}, for each of the eight novel objects. It correctly generates captions that include these novel objects, compared to the baseline model that does not include them.

\subsection{Ablation Studies}
\label{sec:ablation}

In Table~\ref{tab:ablations_diff_steps}, we provide the test split results from our extensive ablation studies on different components of our PS-NOC approach. 

We denote the three Steps in our approach as I, II and III respectively. II denotes training using cross-entropy loss, while II(SCST) and II(SCST-F1) denote additional training using regular CIDEr based SCST and our SCST-F1 respectively. III denotes training using cross-entropy loss, while III(SCST) and III(SCST-F1) denote training using regular SCST and our SCST-F1 instead. We also run a trial with CBS decoding at inference on top of Step I, and denote it as `I + CBSInf'. 
This Table demonstrates the following.

\textbf{i.} Using our synthetic data improves the out-of-domain scores over the baseline, as shown in Row 1 vs. Row 3. It also provides higher scores than running CBS only at inference, as shown in Row 2 vs. Row 3. 

\textbf{ii.} Using our pseudo-labeling approach on top of Step II improves the out-of-domain scores, as shown in all the rows with an additional `+ III'; for e.g., Row 4 vs. Row 3. 

\textbf{iii.} Using SCST during Steps II or III improves the out-of-domain scores over the ones with cross-entropy loss training, as shown in all rows with an additional `(SCST)'; for e.g., Row 5 vs. Row 3. 

\textbf{iv.} Using SCST-F1 during Steps II or III improves the out-of-domain scores over using regular SCST, as shown in all rows with additional `-F1'; for e.g., Row 8 vs. Row 5, and Row 10 vs. Row 7. 

\textbf{v.} Finally, Row 10 and Row 11 are the results from our complete PS-NOC approach, and these have the highest scores. This is an improvement of 85.9 F1-score and 34.1 CIDEr points over the baseline model (Row 1 vs. Row 11) that does not leverage partially paired data.

Additional ablation studies are included in the Supplemental material.

\section{Limitations and Future Work}
In this section, we discuss the limitations of our work and the potential future work to address them.

\textbf{Object annotations for fully paired objects}: During synthetic data generation, our approach requires that fully paired dataset includes bounding box annotations for objects, which is true for both the existing novel captioning datasets~\cite{hendricks2016deep, agrawal2019nocaps} based on MS COCO. However, if such annotations are not available, this requirement can be relaxed by predicting these object bounding boxes using object detection models, and then using them to generate synthetic images.

\textbf{Noise in synthetic data}:
There could be cases where ground truth object bounding box annotations are incorrect or partial, which would result in noisy synthetic images. There could also be cases where our simple caption processing heuristics fail, or the context of synthetic data may not be completely accurate, resulting in noisy synthetic captions. For e.g., `a fraction of itself cut and eaten' is not accurate description of the synthetic `pizza' in Figure~\ref{fig:aug_examples}. 
However, our empirical results (Row 1 vs. 3 in Table~\ref{tab:ablations_diff_steps}) show that synthetic data generated using such simple heuristics is also able to considerably improve the NOC model results. More sophisticated techniques to remove such noise could improve the captioning results further. 

\textbf{Other datasets and models}:
We focused our extensive evaluation only on Up-Down captioning model and \textit{held-out} MS COCO dataset. Evaluation of PS-NOC on the newer Transformer based captioning models~\cite{hu2020vivo} and \textit{no-caps} dataset~\cite{agrawal2019nocaps} will be pursued as future work.

\section{Conclusion}
In this work, we proposed a model-agnostic Partially-Supervised Novel Object Captioning solution, PS-NOC. PS-NOC primarily focuses on the training approach that uses fully paired image-caption data and partially paired images with object detection labels and bounding box annotations. We first generate paired synthetic data within the context of fully paired data. Since, we generate synthetic images instead of region features, our approach is also applicable to the newer captioning models that use the entire image as inputs. We then propose a training approach to train the captioning model for novel objects using this additional synthetic data and partially paired data. This includes novel training techniques such as SCST-F1 and offline pseudo-labeling for NOC. 
Using UpDown captioning model as baseline, our PS-NOC approach sets new state-of-the-art results on \textit{held-out} MS COCO out-of-domain test split with 85.9 F1-score and 103.8 CIDEr. This is an improvement of 85.9 F1-score and 34.1 CIDEr points over the baseline model that does not leverage partially paired data. 


{\small
\bibliographystyle{ieee_fullname}
\bibliography{psnoc}
}

\clearpage
\section{Supplemental material}

\begin{table*}[t]
\caption{The top similar candidate objects in fully paired data for each novel object class. This list is filtered down from the list in~\cite{cao2020feature}} 
\vspace{-4mm}
\label{tab:sim_objs}
\begin{center}
\begin{small}
\begin{tabular}{cccccccc}
\toprule
bottle &  bus &  couch &  microwave &  pizza &  racket &  suitcase &  zebra \\
\midrule
cup &  truck &  chair & refrigerator  &  sandwich &  bat &  handbag &  giraffe \\
wine glass &  car &  bed & toaster  & cake  &  frisbee &  backpack &  elephant \\
vase &  train &  bench &   &   &   &   &  cow \\
\bottomrule
\end{tabular}
\end{small}
\end{center}
\end{table*}

\begin{table*}[!t]
\vskip -0.1in
\caption{\small Comparison of our PS-NOC baseline Step I results against PS3 baseline results on validation split.  Using ResNeXt-152 image features does not give us an unfair advantage while comparing our PS-NOC results with previous works or state-of-the-art results, since our out-of-domain CIDEr and F1 scores are comparable to PS3 baseline scores. We later fine-tune the model from Row 2 here using PS-NOC Steps II and III.} 
\vspace{-6mm}
\label{tab:baslineval}
\vskip 0.15in
\begin{center}
\begin{small}
\begin{tabular}{cccccccccccc}
\toprule
&&&&\multicolumn{4}{c}{Out-of-domain} && \multicolumn{3}{c}{In-domain} \\\cmidrule{5-8}\cmidrule{10-12}
Row & Approach & Training data &&  S &  M &  C &  F1 &&  S &  M &  C \\\midrule
1 & PS3 & In-domain (i.e., fully paired only) &&  14.4 &  22.1 &  69.5  &  0.0  &&  19.9 &  26.5 &  108.6 \\
2 & Ours & In-domain (i.e., fully paired only) &&  19.4 &  27.9 &  66.7  &  0.0  &&  19.5 &  27.8 &  107.8 \\
3 & PS3 & In-domain + Out-of-domain &&  20.1 &  27 &  111.5   &  69.0 &&  20.0 &  26.7 &  109.5 \\
4 & Ours & In-domain + Out-of-domain &&  20.6 &  29.1 &  107.1 &  60.0 &&  20.7 &  29   &  109.0 \\
 \bottomrule
\end{tabular}
\end{small}
\end{center}
\vskip -0.1in
\end{table*}


\subsection{Synthetic Data Generation}
For each novel object, we use the top similar candidate objects that were identified by FDM-net~\cite{cao2020feature}. For e.g., images with `cup', `glass', or `vase' are considered as candidates to generate synthetic data for novel object `bottle'.  We exclude four of these candidate object classes (`stove', `bread', `box', and `racquet') as they do not have bounding box annotations in MS COCO dataset. We show our final list in Table~\ref{tab:sim_objs}.

\subsection{Baseline Results}

In order to get a good baseline model for Step I of our PS-NOC approach, we train the Up-Down model~\cite{anderson2018bottom} using only the in-domain fully paired data and compare the validation split CIDEr against the baseline results of PS3~\cite{anderson2018partially}, which also uses the Up-Down model. With image features extracted using our ResNet-101 backbone based Faster R-CNN, our Up-Down model baseline under-performed compared to PS3 baseline. We were however able to achieve comparable baseline results by using ResNeXt-152 backbone based image features, as shown in Row 1 vs. Row 2 of Table~\ref{tab:baslineval}. 
We also train another baseline model on the complete MS COCO dataset (i.e. 82,783 image-caption pairs), and compare our results against the corresponding baseline results in PS3; these are shown in Row 3 vs. Row 4 of Table~\ref{tab:baslineval}. 

Both these results show that our primary metrics of comparison, CIDEr and F1-score on out-of-domain test set, are either comparable or slightly lower than the baseline results in PS3, and that the ResNeXt-152 image features do not provide us an unfair advantage during comparison of PS-NOC with previous works or other state-of-the-art results.


\begin{table*}[!t]
\caption{\small Test split results of our ablation study showing the benefits of using i. Our pseudo-labeling approach, and ii. More synthetic data.} 
\vspace{-4mm}
\label{tab:ablations_diff_steps_supp}
\begin{center}
\begin{small}
\begin{tabular}{ccccccccccccc}
\toprule
& &\multicolumn{4}{c}{Out-of-domain} & & \multicolumn{3}{c}{In-domain} \\\cmidrule{3-6}\cmidrule{8-10}
Row & Training technique &  S &  M &  C &  F1 &&  S &  M &  C \\\midrule
1 &  I + III(CBS) &  19.9 &  28.1 &  92.7 &  60.8 &&  19.6 &  27.8 &  105.9 \\
2 &  I + III &  19.4 &  28.0 &  86.9 &  60.9 &&  19.1 &  27.6 &  105.5 \\
3 &  I + II(33\%) + III(CBS)  &  20.2 &  28.3 &  92.4 &  68.8 &&  19.8 &  27.9 &  106.7 \\
4 &  I + II(33\%) + III &  20.1 &  28.0 &  93.7 &  69.1 &&  19.7 &  27.7 &  105.6 \\
5 &  I + II + III(CBS) &  20.0 &  28.3 &  94.4 &  74.1 &&  19.6 &  27.6 &  105.0 \\
6 &  I + II + III  &  20.1 &  28.3 &  96.5 &  74.0 &&  19.7 &  27.8 &  105.0 \\
 \bottomrule
\end{tabular}
\end{small}
\end{center}
\end{table*}

\subsection{Ablation Studies}
\label{sec:ablation}

We perform ablation study to evaluate the effectiveness of our pseudo-labeling approach (see  Section~\ref{sec:training_tech}), wherein we generate two pseudo-label captions per partially paired image, one caption using CBS and the other without CBS. We also study the impact of the volume of synthetic data used during our PS-NOC training on the captioning results.

In Table~\ref{tab:ablations_diff_steps_supp}, we provide the test split results from these additional ablation studies. We denote the three Steps in our PS-NOC approach as I, II and III respectively. II denotes cross-entropy loss training using our entire synthetic data (i.e., $K = 2400$ images per novel object), while II(33\%) denotes similar training done using only 33\% of this synthetic data (i.e., $K = 800$ images per novel object). III denotes cross-entropy loss training done using our pseudo-labeling approach, while III(CBS) denotes similar cross-entropy loss training but done using only the CBS generated pseudo-label caption. In these studies, we follow the same training schedule for Steps I and II as in Section~\ref{sec:imp_details}. In Step III, we train the model for 15,000 iterations per round with an initial LR of 0.0025 for the first round and scale it by 0.5 across rounds.
Table~\ref{tab:ablations_diff_steps_supp} demonstrates the following.

\textbf{i.} Our pseudo-labeling approach improves the out-of-domain caption quality, i.e., CIDEr scores, as seen in Row 3 vs. Row 4 and Row 5 vs. Row 6. This improvement is only seen if we use synthetic data (Rows 3-6) as proposed in our PS-NOC. (Rows 1 and 2 do not use synthetic data for training, as denoted by the missing II,)  

\textbf{ii.} Using more synthetic data with our PS-NOC approach helps improve the out-of-domain scores as seen in Row 3 vs. Row 5 and Row 4 vs. Row 6.






\begin{table*}[t]
\caption{\small Using synthetic data in addition to complete MS COCO training data improves out-of-domain CIDEr and F1 scores on test split.}
\vspace{-4mm}
\label{tab:ablations_syn_data}
\begin{center}
\begin{small}
\begin{tabular}{ccccccccc}
\toprule
&\multicolumn{4}{c}{Out-of-domain} & & \multicolumn{3}{c}{In-domain}  \\\cmidrule{2-5}\cmidrule{7-9}
Additional Synthetic Training data &  S &  M &  C &  F1 &&  S &  M &  C  \\
 \midrule
No &  21.0 &  29.4 &  110.2 &  60.8 &&  20.6 &  29.1 &  109.6 \\
Yes &   20.8 &  28.8 &  \textbf{111.3} &  \textbf{65.6} &&  20.5 &  28.9 &  108.3  \\ 
\bottomrule
\end{tabular}
\end{small}
\end{center}
\end{table*}

\subsection{Caption Post-processing and Soln-2}
\begin{figure*}[!ht]
\captionsetup[subfigure]{labelformat=empty}
		\centering     
		
		\begin{subfigure}[t]{0.23\textwidth}
		\caption{zebra}
		\end{subfigure}
		\hspace{1mm}
		\begin{subfigure}[t]{0.23\textwidth}
		\caption{bus}
		\end{subfigure}
		\hspace{1mm}
		\begin{subfigure}[t]{0.23\textwidth}
		\caption{couch}
		\end{subfigure}
		\hspace{1mm}
		\begin{subfigure}[t]{0.23\textwidth}
		\caption{microwave}
		\end{subfigure}		
		
		\begin{subfigure}[t]{0.23\textwidth}
		
		\includegraphics[width=1.0\textwidth, height=28mm]{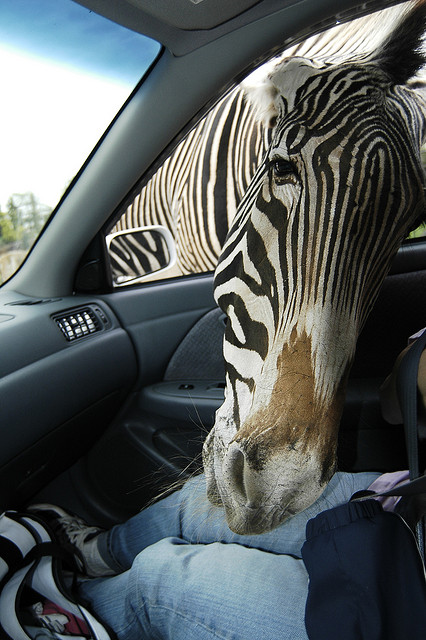}
		\caption{\textbf{Baseline}: A person sitting in a car with a large giraffe. }		
	
		\end{subfigure}
		\hspace{1mm}
		\begin{subfigure}[t]{0.23\textwidth}
		\includegraphics[width=1.0\textwidth, height=28mm]{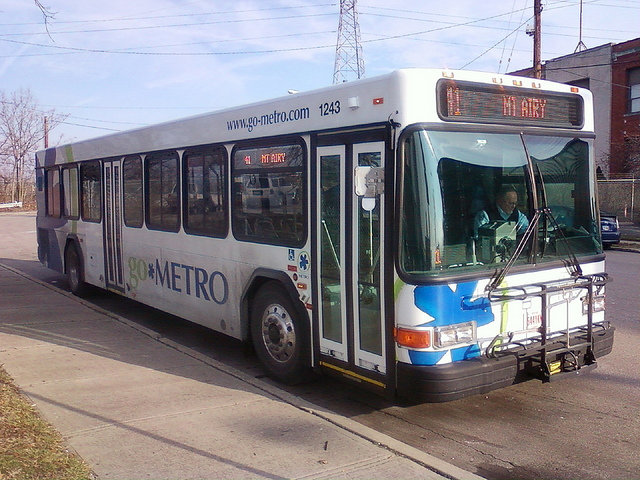}
		\caption{\textbf{Baseline}: A police car parked on the side of the road.  
		}
		\end{subfigure}
		\hspace{1mm}
		\begin{subfigure}[t]{0.23\textwidth}
		\includegraphics[width=1.0\textwidth, height=28mm]{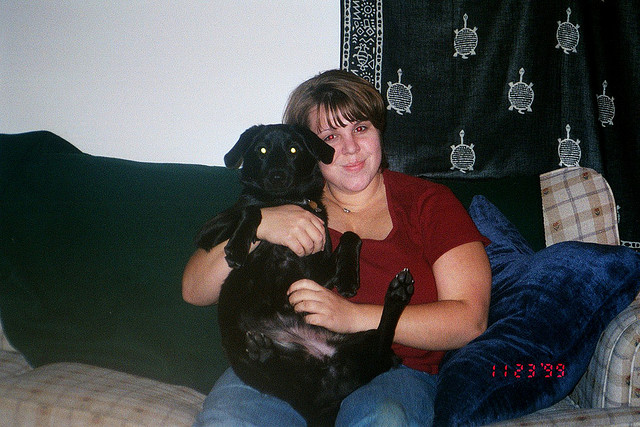}
		\caption{\textbf{Baseline}: A woman is holding a black dog and a black dog.	}
		
		\end{subfigure}
		\hspace{1mm}
		\begin{subfigure}[t]{0.23\textwidth}
		\includegraphics[width=1.0\textwidth, height=28mm]{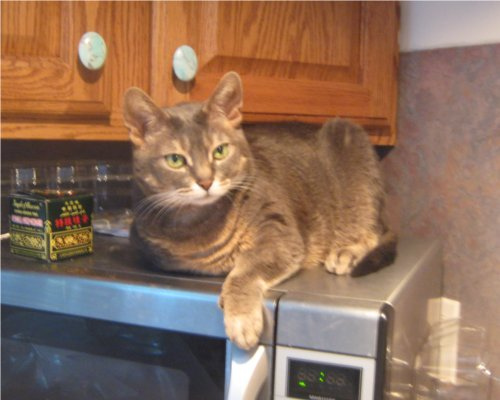}
				\caption{
		\textbf{Baseline}: A cat is sitting on top of a toaster.
		}
		\end{subfigure}

		\begin{subfigure}[t]{0.23\textwidth}
		\caption{
		\textbf{Soln-1}: A \textbf{\underline{zebra}} is standing in a car with a.\\\textbf{Soln-2}: A \textbf{\underline{zebra}} is standing in the seat of a car.}
		\end{subfigure}
		\hspace{1mm}
		\begin{subfigure}[t]{0.23\textwidth}
		\caption{
		\textbf{Soln-1}: A \textbf{\underline{bus}} is parked on the side of a.\\\textbf{Soln-2}: A \textbf{\underline{bus}} is parked on a city street.}
		\end{subfigure}
		\hspace{1mm}
		\begin{subfigure}[t]{0.23\textwidth}
		\caption{\textbf{Soln-1}: A woman sitting on a \textbf{\underline{couch}} with a black dog in.\\\textbf{Soln-2}: A woman sitting on a \textbf{\underline{couch}} holding a black dog.}
		\end{subfigure}
		\hspace{1mm}
		\begin{subfigure}[t]{0.23\textwidth}
		\caption{
		\textbf{Soln-1}: A cat laying on top of a \textbf{\underline{microwave}} in a.\\\textbf{Soln-2}: A cat laying on top of a \textbf{\underline{microwave}}.}
		\end{subfigure}

		\vspace{2mm}
		
		\begin{subfigure}[t]{0.23\textwidth}
		\caption{pizza}
		\end{subfigure}
		\hspace{1mm}
		\begin{subfigure}[t]{0.23\textwidth}
		\caption{racket}
		\end{subfigure}
		\hspace{1mm}
		\begin{subfigure}[t]{0.23\textwidth}
		\caption{suitcase}
		\end{subfigure}
		\hspace{1mm}
		\begin{subfigure}[t]{0.23\textwidth}
		\caption{bottle}
		\end{subfigure}
		
		\begin{subfigure}[t]{0.23\textwidth}
		\includegraphics[width=1.0\textwidth, height=28mm]{figs/sol1_sol2/pizza_COCO_val2014_000000543065.jpg}
		\caption{\textbf{Baseline}: A man sitting at a table with plates of food. }		
	
		\end{subfigure}
		\hspace{1mm}
		\begin{subfigure}[t]{0.23\textwidth}
		\includegraphics[width=1.0\textwidth, height=28mm]{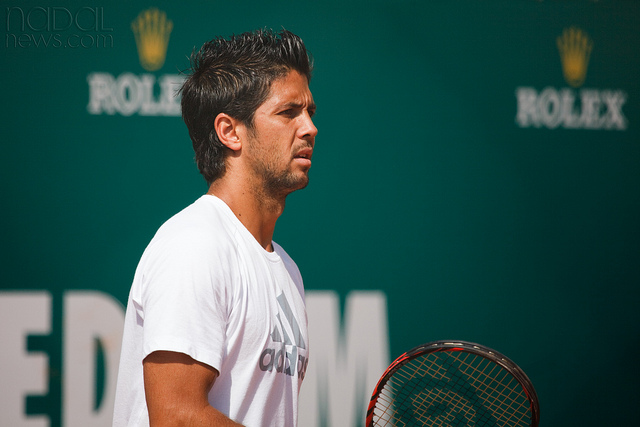}
		\caption{\textbf{Baseline}: A man is standing next to a tennis ball.
		}
		\end{subfigure}
		\hspace{1mm}
		\begin{subfigure}[t]{0.23\textwidth}
		\includegraphics[width=1.0\textwidth, height=28mm]{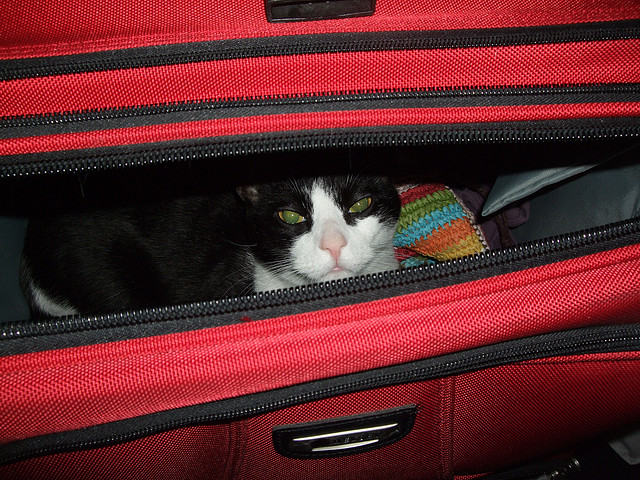}
		\caption{\textbf{Baseline}: A black and white cat is sitting in a car. }
		
		\end{subfigure}
		\hspace{1mm}
		\begin{subfigure}[t]{0.23\textwidth}
		\includegraphics[width=1.0\textwidth, height=28mm]{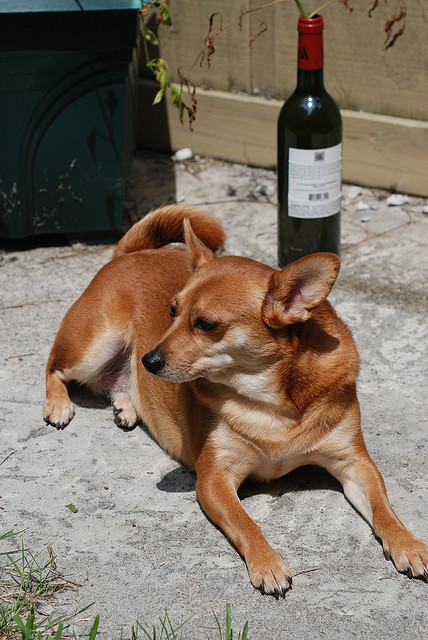}
				\caption{
		\textbf{Baseline}: A brown dog laying next to a glass of wine.
		}
		\end{subfigure}
		
		\begin{subfigure}[t]{0.23\textwidth}
		\caption{
		\textbf{Soln-1}: A man sitting at a table with a \textbf{\underline{pizza}} and.\\\textbf{Soln-2}: A man sitting at a table with a \textbf{\underline{pizza}}.}
		\end{subfigure}
		\hspace{1mm}
		\begin{subfigure}[t]{0.23\textwidth}
		\caption{
		\textbf{Soln-1}: A man holding a tennis \textbf{\underline{racket}}  in front of a.\\\textbf{Soln-2}: A man holding a tennis \textbf{\underline{racket}} on a field.}
		\end{subfigure}
		\hspace{1mm}
		\begin{subfigure}[t]{0.23\textwidth}
		\caption{\textbf{Soln-1}: A black and white cat laying in a \textbf{\underline{suitcase}} on.\\\textbf{Soln-2}: A black and white cat laying in a \textbf{\underline{suitcase}}.}
		\end{subfigure}
		\hspace{1mm}
		\begin{subfigure}[t]{0.23\textwidth}
		\caption{
		\textbf{Soln-1}: A dog laying next to a \textbf{\underline{bottle}} of wine on.\\\textbf{Soln-2}: A dog laying next to a \textbf{\underline{bottle}} of wine.}
		\end{subfigure}
		\caption{Examples of captions generated using our PS-NOC approach compared to the baseline model for the eight novel object classes. Our Soln-1 and Soln-2 both correctly include the novel object classes in the captions. Soln-2 includes caption post-processing during training and inference and has better caption quality than our Soln-1.}
\label{fig:sol1_sol2_cap_examples}	
\end{figure*}

We notice that using SCST during training sometimes results in captions that end with the words `in', `a', `with', etc. To improve caption quality, we perform post-processing during SCST-F1 training and inference, and remove such words at the end of the caption. We refer to our approach without such post-processing as Soln-1, and the one with the post-processing as Soln-2. 
The complete list of these words is as follows: 'with','in','on','of','a','at','to','for','an','this','his','her','that', 'the'.
We provided the results for both these solutions in Table~\ref{tab:results}. 
We show a few qualitative examples of these results in Figure~\ref{fig:sol1_sol2_cap_examples}.

We used a slightly different training schedule for PS-NOC Step III in our Soln-2, compared to our Soln-1.
In our Soln-1, we fine-tuned the model using SCST-F1 for $N=4$ rounds with 6,000 iterations per round, initial LR of 0.002 for first round and scale it by 0.8 across rounds. 
In our Soln-2, we fine-tuned the model using SCST-F1 for $N=4$ rounds with 8,000 iterations per round, initial LR of 0.003 for first round and scale it by 0.6 across rounds.

\subsection{Synthetic Data in Addition to Complete Data}
We also check whether our synthetic data is useful to captioning models that have access to the complete training data, i.e., regular image captioning problem (not NOC). 

We run a trial where we train the Up-Down model from scratch using both the complete MS COCO training data (i.e. 82,783 image-caption pairs) and our synthetic data (i.e. 18,974 image-caption pairs), and compare it against the model trained without using synthetic data. We train both these models for similar number of epochs, i.e., 49,000  and 40,000 iterations respectively to account for their dataset size difference, using a batch size of 100 and same LR schedule. The results in Table~\ref{tab:ablations_syn_data} on the test split show that using this additional synthetic data improves both out-of-domain CIDEr and F1-score by 1 and 5 points respectively. This encourages us to invest future research effort in generating such synthetic data for in-domain objects as well. 

\end{document}